\documentclass[runningheads]{llncs}

\usepackage{ifpdf}

\usepackage{amssymb}
\usepackage{graphicx}
\usepackage{url}
\usepackage{enumerate}
\usepackage{subfigure}
\usepackage{amsmath}
\usepackage[linesnumbered,boxed,ruled]{algorithm2e}

\usepackage{multirow}
\usepackage{cases}

\newcommand{\fref}[1]{Fig.~\ref{#1}}
\newcommand{\sref}[1]{Section \ref{#1}}

\newcommand{\eref}[1]{Eq.~(\ref{#1})}

\begin{document}
\title{Investigating Constraint Relationship in Evolutionary Many-Constraint Optimization\thanks{This work is supported by the National Natural Science Foundation of China (Nos. 61773390 and 72071205), the Hunan Youth elite program (2018RS3081), the scientific key research project of National University of Defense Technology (ZK18-02-09, ZZKY-ZX-11-04) and the key project of 193-A11-101-03-01.}}
%
%
%
\titlerunning{Investigating Constraint Relationship in EMCO} 
\author{Mengjun Ming$^{1,2}$\and Rui Wang$^{1,2}$ \and Tao Zhang$^{1,2}$}
\authorrunning{M. Ming et al.} 

\institute{
	1:~College of Systems Engineering, National University of Defense Technology, Changsha 410073, China
	\\2:~Hunan Key Laboratory of Multi-Energy System Intelligent Interconnection Technology (HKL-MESI$^2$T), Changsha 410073, China 
\\ Correspondence: \email{ruiwangnudt@gmail.com}}

\maketitle

\begin{abstract}
This paper contributes to the treatment of extensive constraints in evolutionary many-constraint optimization through consideration of the relationships between pair-wise constraints. In a conflicting relationship,  the functional value of one constraint increases as the value in another constraint decreases. In a harmonious relationship, the improvement in one constraint is rewarded with simultaneous improvement in the other constraint. In an independent relationship, the adjustment to one constraint never affects the adjustment to the other. Based on the different features, methods for identifying constraint relationships are discussed, helping to simplify many-constraint optimization problems (MCOPs). Additionally, the transitivity of the relationships is further discussed at the aim of determining the relationship in a new pair of constraints.
\end{abstract}

\keywords{
Many-constraint optimization problems (MCOPs) \and Constraint relationships \and Conflict \and Harmony \and Independence.
}

\section{Introduction}

Constrained optimization problems (COPs) refer to finding solutions that satisfy all constraints and optimize the objective functions~\cite{Coello2002Constraint}. In literature numerous methods have been developed to handle COPs. However, the concerned COPs often has only a small number of constraints, which thus, cannot adapt to real-world problems since they often involve in a large number of constraints. Within a heavily constrained space, searching for optimal solution(s) becomes increasingly challenging~\cite{fan2019difficulty}. Moreover, constraints often feature differently in terms of the handling difficulty and/or the relative importance~\cite{zhang2019soft}, in addition to the preference information from decision-makers. As a consequence, a novel question emerges, that is, how to effectively handle large number of constraints given to their different features. 

With respect to the above issue, the primary concern is to analyze the relationships between constraints. A good understanding of the relationships can certainly facilitate the design of constraint-handling strategies. However, qualitative studies of pair-wise constraint relationships are not common in neither academia nor industry. Motived by this, this study investigates the possible relationships amongst constraints. For a clear description,  in this study the optimization problem with many different constraints is termed many-constraint optimization problem (MCOP). Following that, the types of constraints are systematically analyzed. In specific, ``conflict'', ``harmony'' and ``independence'' relationships are mainly discussed. 

The rest of this study is organized as follows. \sref{sec:MCOP} formulates MCOPs and introduces the classification of possible constraint relationships. The conflicting relationship between constraints is discussed in \sref{sec:Conflict} whilst \sref{sec:Harmony} considers harmonious constraints. And the case where constraints are independent of each other is presented in \sref{sec:Independence}. Then methods to identify pair-wise relationships are described in \sref{sec:Method}, followed by the discussion about the transitivity between constraint relationships in \sref{sec:Transitivity}. Finally, \sref{sec:conclusion} concludes the paper and identifies future studies.

\section{Problem Formulation and Motivation}
\label{sec:MCOP}
 
Without loss of generality, a constrained optimization problem can be formulated as follows (for minimization)~\cite{Cai2006A}:
 \begin{equation}
\begin{gathered}
\min f(\mathbf{x}),\mathbf{x}=(x_1, x_2, \ldots ,x_n),
\\
\end{gathered}
\label{eq:COP}
\end{equation}
where $\mathbf{x}$ is a decision vector in the $n$-dimensional space (i.e. $\mathbf{x}\in \Re ^n$) and $f(\cdot)$ is the objective function. Each element in $\mathbf{x}$ (e.g., $x_k$), is bounded by the lower and upper limits (e.g., $\underline{x}_k$ and $\overline{x}_k$, respectively), as shown in \eref{eq:variable}.
\begin{equation}
\begin{gathered}
\underline{x}_k\leq x_k\leq \overline{x}_k,1\leq k\leq n.
\\
\end{gathered}
\label{eq:variable}
\end{equation}
Additionally, there are often a set of $m(m\geq 0$) linear or nonlinear constraints in the search space: 
\begin{equation}
\begin{gathered}
g_j(\mathbf{x})\leq 0,j = 1,\ldots ,q,\\
h_j(\mathbf{x})=0,j=q+1,\ldots, m,
\\
\end{gathered}
\label{eq:constraint}
\end{equation}
where $q$ and $m-q$ are the number of inequality and equality constraints, respectively. 
When the number of constraints is large, this kind of problem is termed \textbf{many-constraint optimization problems (MCOPs)}~\cite{ming2019evolutionary}.

We argue that MCOPs should be paid more attention since many real-world problems involve hundreds of or even more constraints.
They are evidently distinct from the existing constrained problems because their excessive constraints bring them new challenges. The first demand is to sort the constraints out. Involved in such intricate constraints, it is unreasonable to hastily consider all of them as a whole without priority analysis. Instead, the appropriate distinction and grouping based on their inner relationships can make some sense, and thus the pair-wise relationships between constraints are investigated in this paper. 

%

\section{Relationships between Constraints}
\label{sec:Relationship}
Among a large number of constraints, different relationships may exist between two constraints. Before elaborating these relationships, we firstly introduce some notations.

Let $j$ be the index to a particular constraint: $j\in \left\lbrace 1,2,\ldots,m\right\rbrace$ and  $f_j$ be the functional value of the $j$th constraint. That is, 
\begin{equation}
\begin{gathered}
f_j(\mathbf{x})=
\begin{cases}
g_j(\mathbf{x})&\mbox{$j=1,2,\ldots,q$},\\
h_j(\mathbf{x})&\mbox{$j=q+1,\ldots,m$}.\\
\end{cases}\\
\end{gathered}
\label{eq:MCOP2}
\end{equation}

Though it is widely acknowledged that the attention in constrained optimization problems is whether the constraint is satisfied instead of the satisfying extent, the comparison of $f_j$ between solutions can help us to know the changing tendency of the constraints, which contributes to the analysis of their inner relationships.

\begin{figure}[tb]
\begin{centering}
\includegraphics[width=0.6\textwidth]{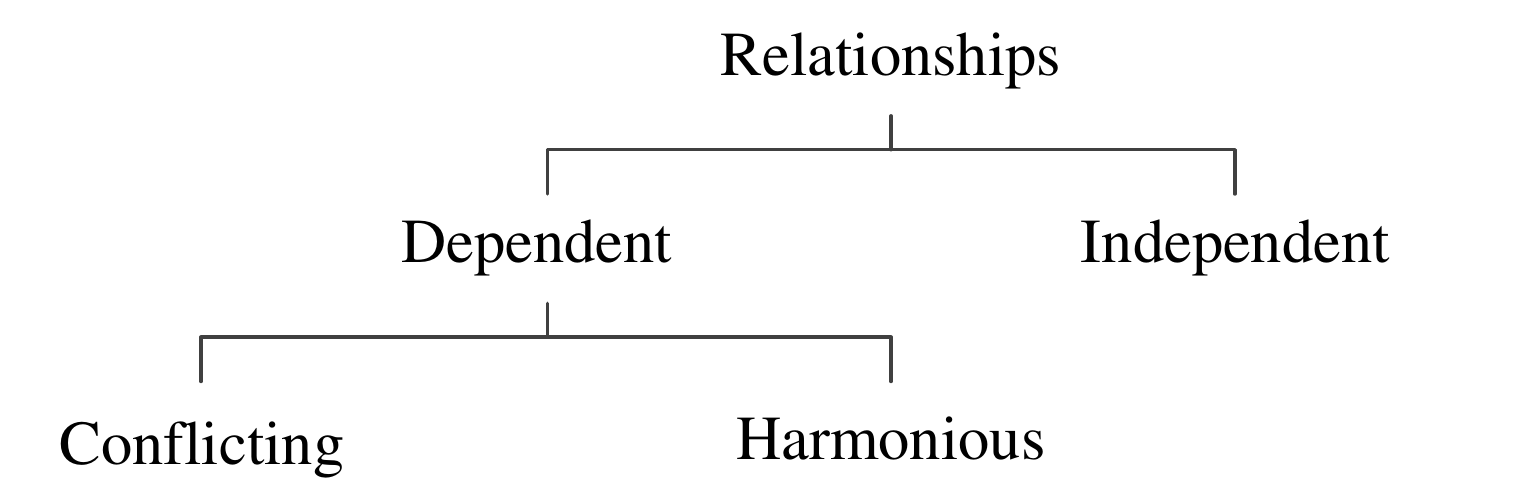}
\par\end{centering}
\caption{Possible relationships between different constraints.}
\label{Fig:Relationship}
\end{figure}

For two constraints, e.g., the $i$th constraint and the $j$th constraint where $i,j\in \left\lbrace 1,2,\ldots,m\right\rbrace$, there are mainly three different cases with respect to the comparison result between their functional values. Correspondingly, the constraint relationships are classified into three types, as shown in \fref{Fig:Relationship}. 

\subsection{Conflicting Constraints}
\label{sec:Conflict}
Similar to the definition of conflicting objectives in~\cite{Purshouse2003Conflict}, we consider the relationship of pair-wise constraints as \textit{conflicting} when the functional value of one constraint decreases as the value in another constraint increases. It can be expressed as follows:
Constraints $i$ and $j$ exhibit evidence of conflict when $f_i(\mathbf{x}^a)<f_i(\mathbf{x}^b)$ and $f_j(\mathbf{x}^a)>f_j(\mathbf{x}^b)$ where $\mathbf{x}^a$ and $\mathbf{x}^b$ are two different individuals. If $\exists (\mathbf{x}^a,\mathbf{x}^b)$ then there is conflict.

\begin{figure}[!t]
\begin{centering}
\subfigure[\label{Fig:Conflict_example1}The contour of the $i$th constraint function ($f_i$)]{\includegraphics[width=0.32\textwidth]{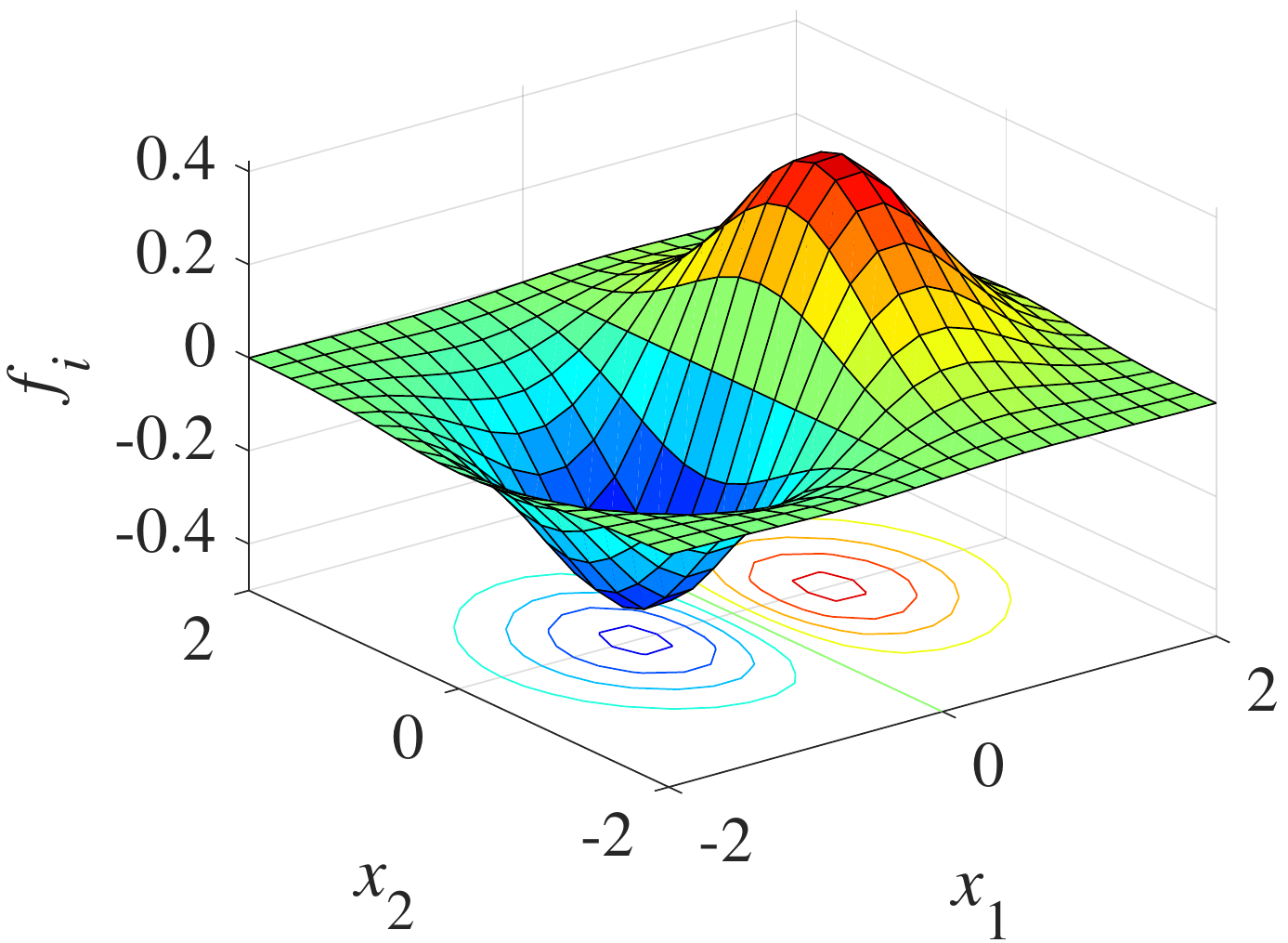}\includegraphics[width=0.07\textwidth]{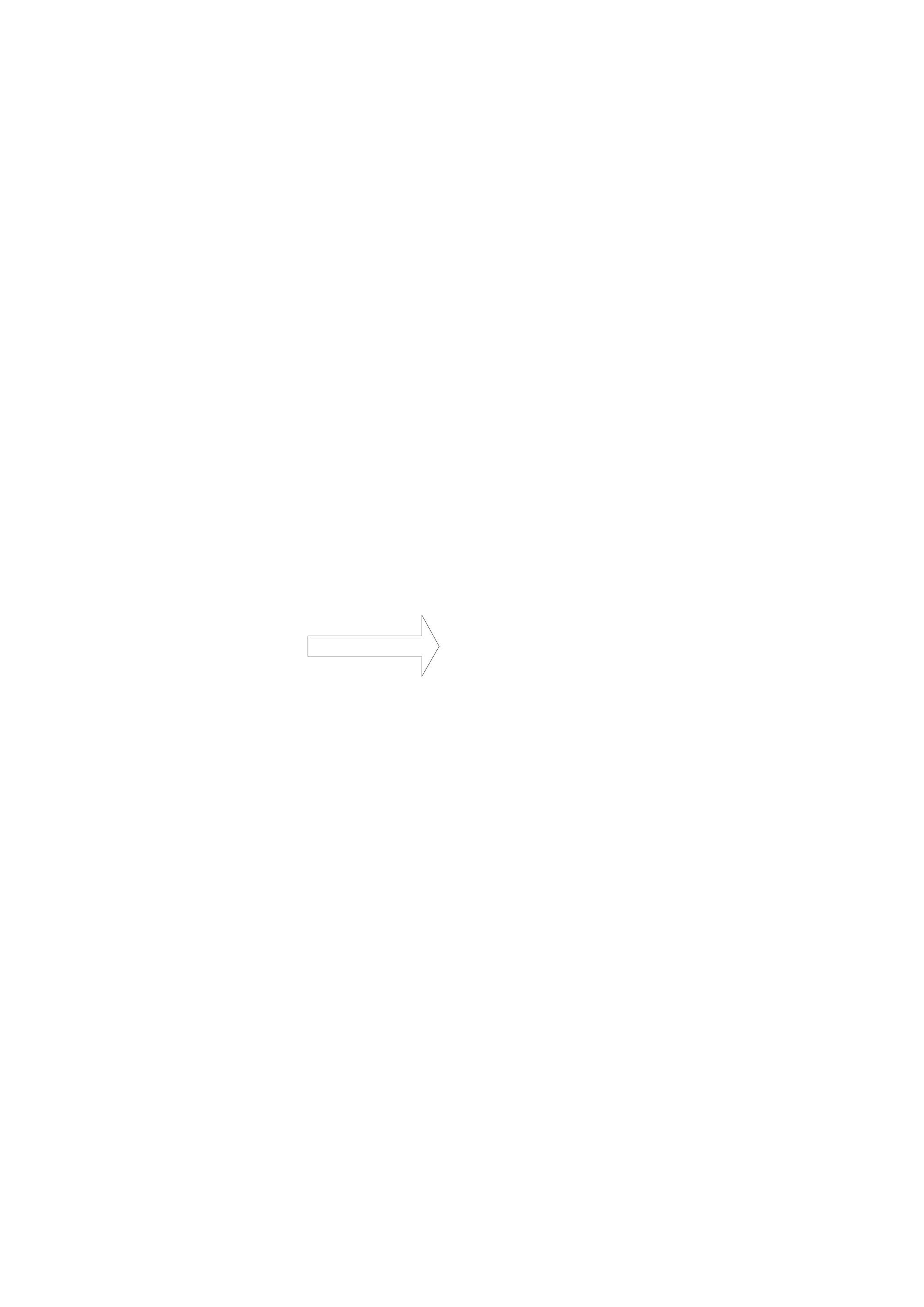}\includegraphics[width=0.32\textwidth]{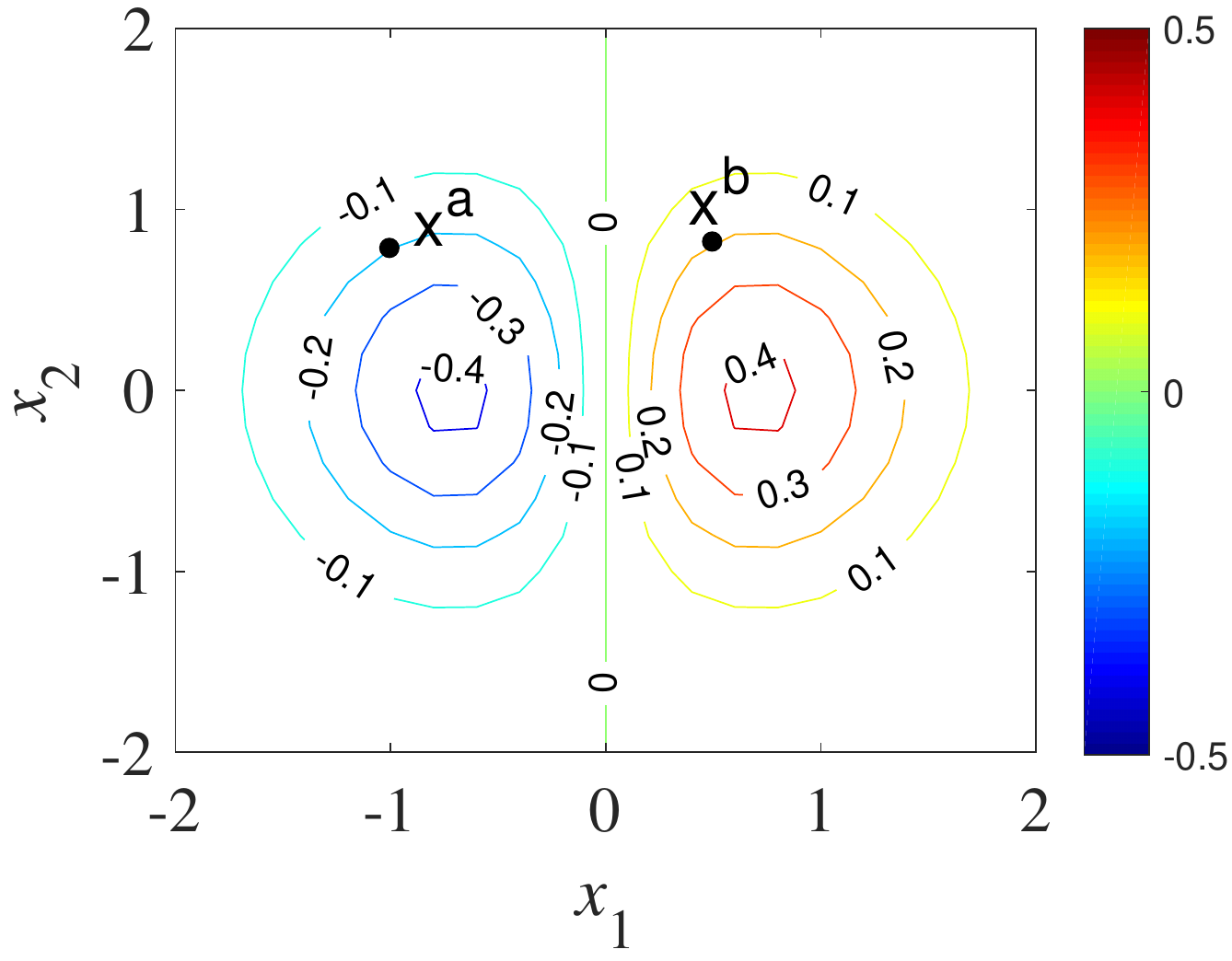}}
\subfigure[\label{Fig:Conflict_example2}The contour of the $j$th constraint function ($f_j$)]{\includegraphics[width=0.32\textwidth]{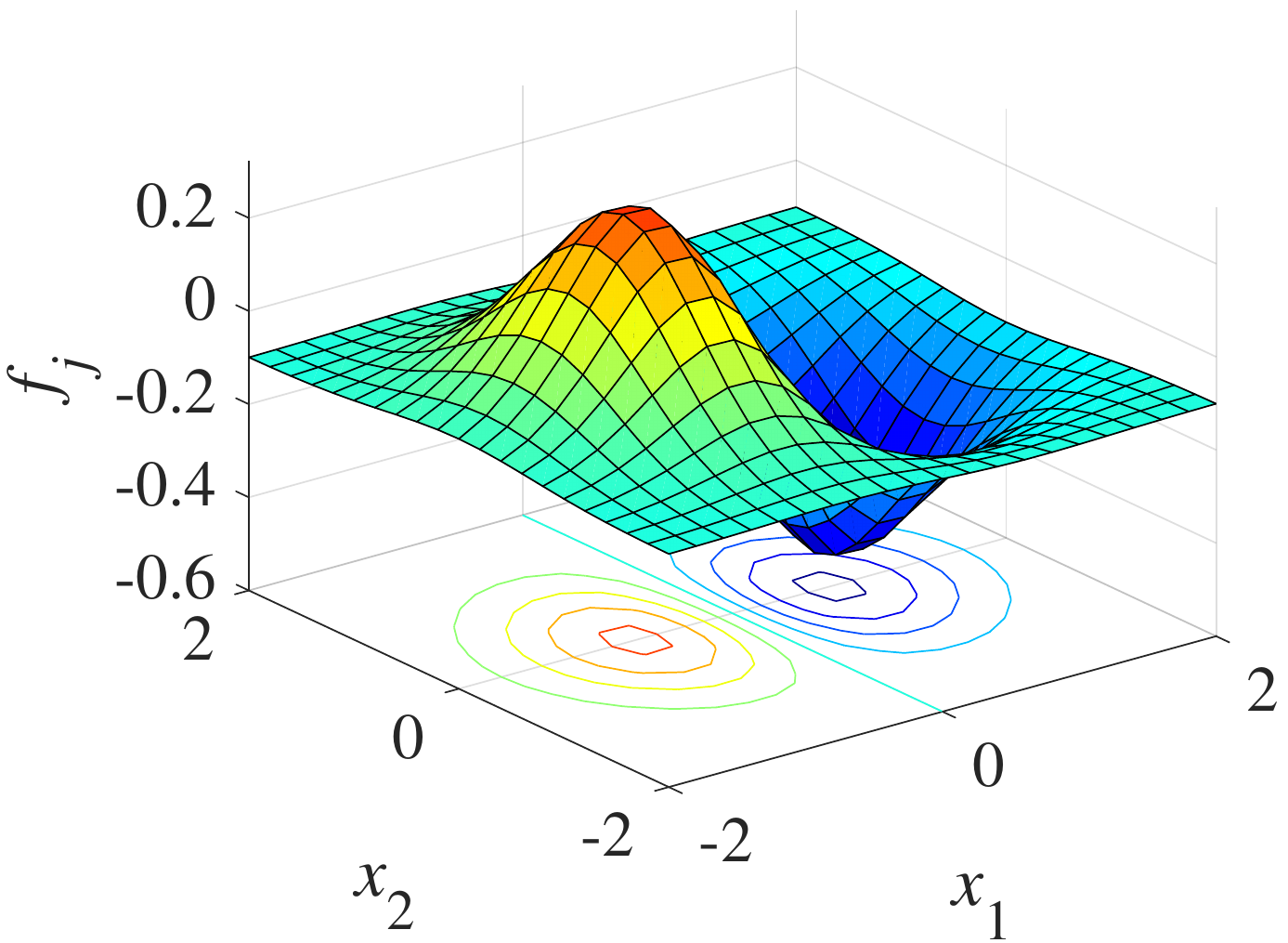}\includegraphics[width=0.07\textwidth]{Arrow}\includegraphics[width=0.32\textwidth]{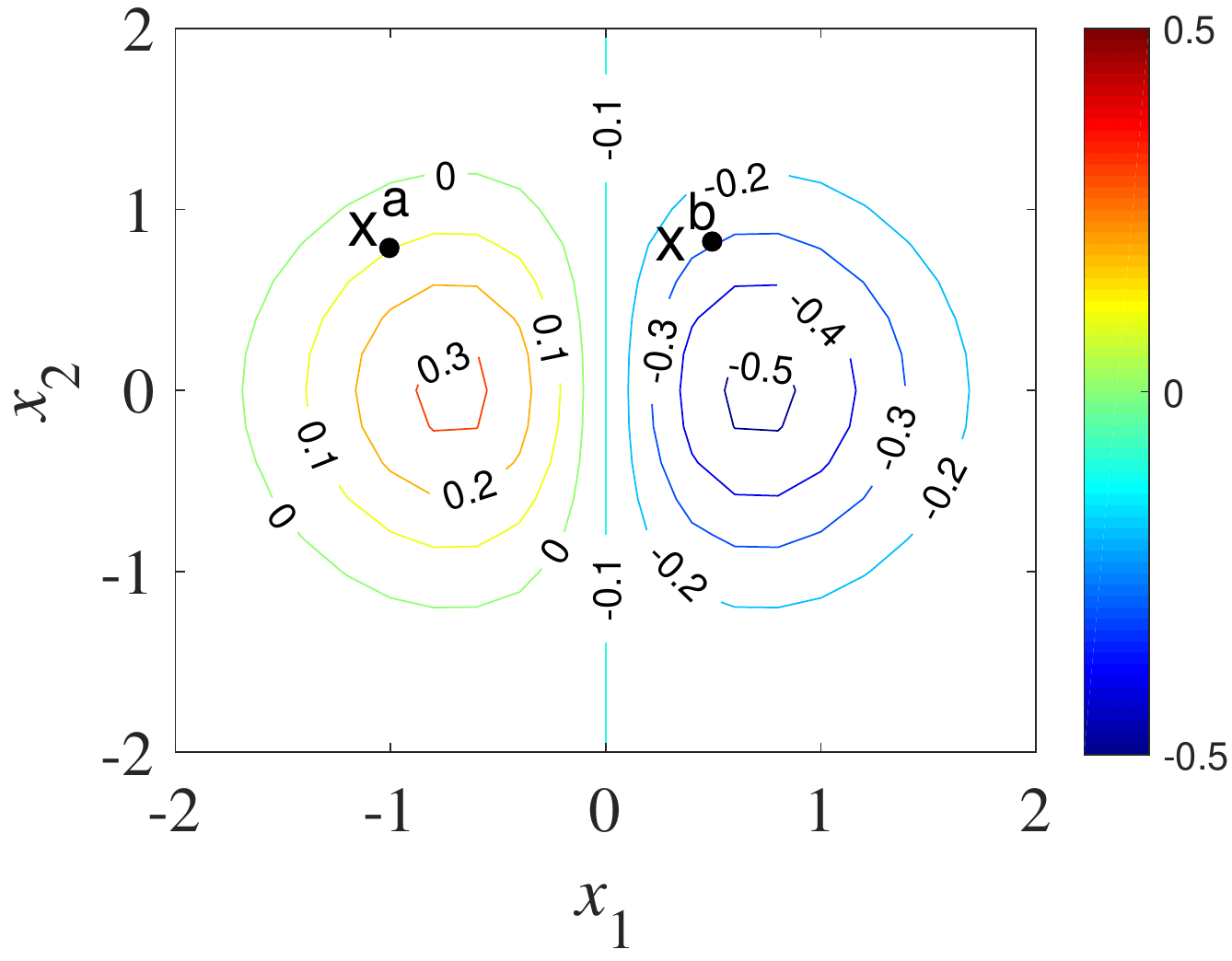}}\\
\subfigure[\label{Fig:Conflict_example}Evidence of conflict between two individuals]{\includegraphics[width=0.6\textwidth]{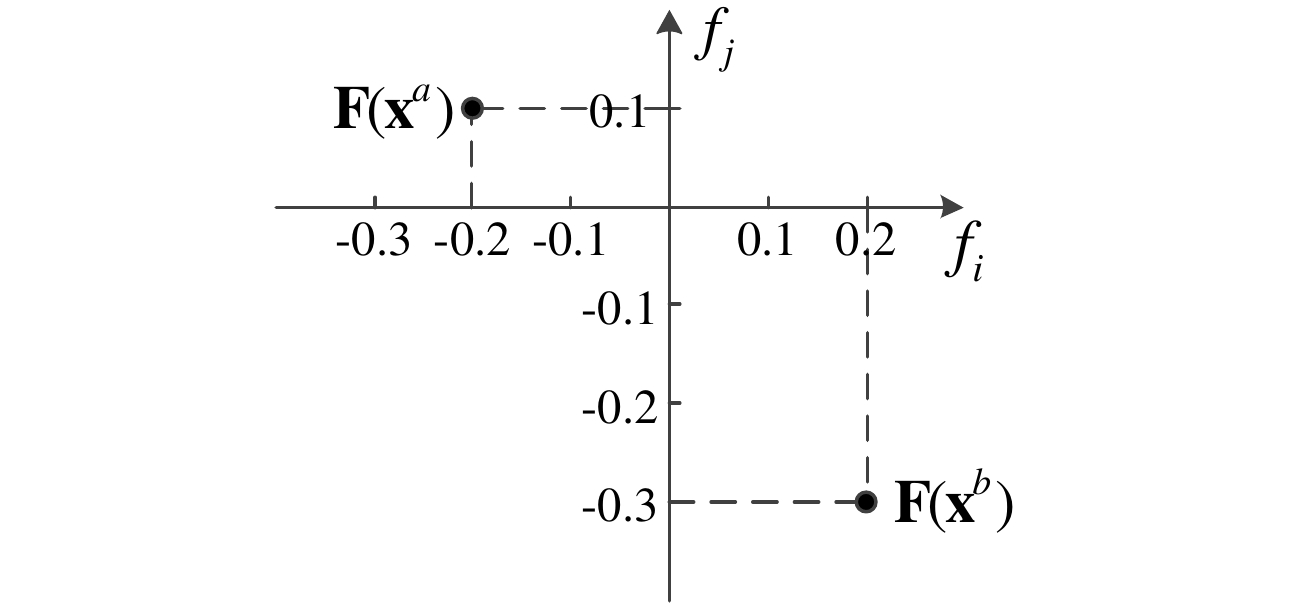}}
\par\end{centering}
\caption{Illustration of two conflicting constraints.}
\label{Fig:Conflict}
\end{figure}

If $\forall(\mathbf{x}^a,\mathbf{x}^b)$ holds the conflicting condition then there is \textit{total conflict}. A simple example of \textit{total conflict} is demonstrated in \fref{Fig:Conflict} where constraints $i$ and $j$ correspond to $x_1e^{-x_1^2-x_2^2}\leq 0$ and $-0.1-x_1e^{-x_1^2-x_2^2}\leq 0$, respectively. It is observed from \fref{Fig:Conflict_example1} and \fref{Fig:Conflict_example2} that the search direction of the two feasible regions is opposite. 
Given any two solutions in the search space (e.g., $\mathbf{x}^a$ and $\mathbf{x}^b$ in \fref{Fig:Conflict}), if $f_i(\mathbf{x}^a)<f_i(\mathbf{x}^b)$ then there is $f_j(\mathbf{x}^a)>f_j(\mathbf{x}^b)$, and vice versa. For instance, \fref{Fig:Conflict_example} shows the evidence of conflict between $\mathbf{x}^a$ and $\mathbf{x}^b$ regarding the functional values of constraints $i$ and $j$. And any other two solutions in the search space also follow this case. Thus the solutions found during the search will form a trade-off surface.

The conflicting condition arises regularly in constrained optimization problems, which causes the difficulty in finding the solutions satisfying the constraints simultaneously, especially when the number of conflicting constraints is large. Specifically, the decrease/increase of the functional value of some constraints may bring the opposite change in other ones. When there are  many conflicting constraints, searching for the optimal solution would become extremely hard due to the push force from different directions.

As mentioned above, large numbers of conflicting constraints can limit the search for feasible regions. In this case, it is not sensible to consider all the constraints as a whole, without differentiating their features. The reason is as follows: in some MCOPs, the proportion of the feasible solutions is pretty small, and when selecting individuals to generate offspring in an early stage, chances are that nearly all solutions are infeasible where how to choose the relatively good ones is a challenging issue.

The exploration of the preferences for the constraints could be an effective strategy to deal with a large number of conflicting constraints. These preferences may be a hard constraint (a requirement that must be met) or a soft aspiration. They can serve as a baseline when comparing those potential infeasible solutions. Specifically, they can be used to build a partial ordering for the constraints, and the solutions satisfying the hard constraints can be preferred.

The preference information may be either qualitative or quantitative, and may also not be available as a priori. No matter in which case, it is desirable to have a promising method to gain the information about the relative importance levels of different constraints.



%
%
%
%

\subsection{Harmonious Constraints}
\label{sec:Harmony}
Here \textit{harmonious} is a relationship where the enhancement of the functional value of a constraint is rewarded with the improvement of another constraint. 
The condition of \textit{harmony} is determined when $f_i(\mathbf{x}^a)<f_i(\mathbf{x}^b)$ and $f_j(\mathbf{x}^a)<f_j(\mathbf{x}^b)$ where $\mathbf{x}^a$ and $\mathbf{x}^b$ are two different individuals. If $\nexists(\mathbf{x}^a,\mathbf{x}^b)$ holds this condition then there is \textit{no harmony}, if $\exists(\mathbf{x}^a,\mathbf{x}^b)$ then there is \textit{harmony}, and if the condition holds for $\forall(\mathbf{x}^a,\mathbf{x}^b)$ then there is \textit{total harmony}.

\begin{figure}[!t]
\begin{centering}
\subfigure[\label{Fig:Harmony_example1}The contour of the $i$th constraint function ($f_i$)]{\includegraphics[width=0.32\textwidth]{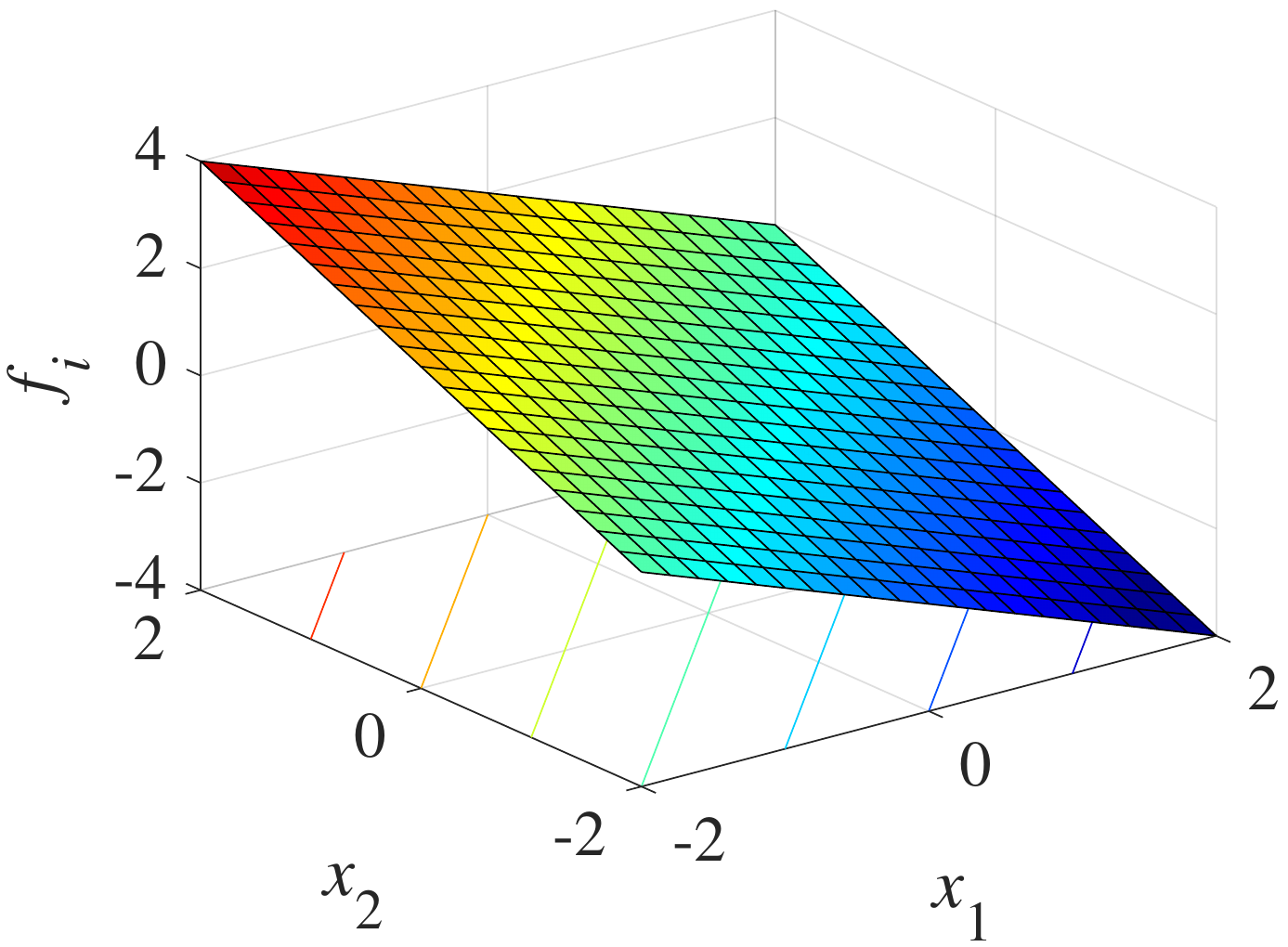}\includegraphics[width=0.07\textwidth]{Arrow}\includegraphics[width=0.32\textwidth]{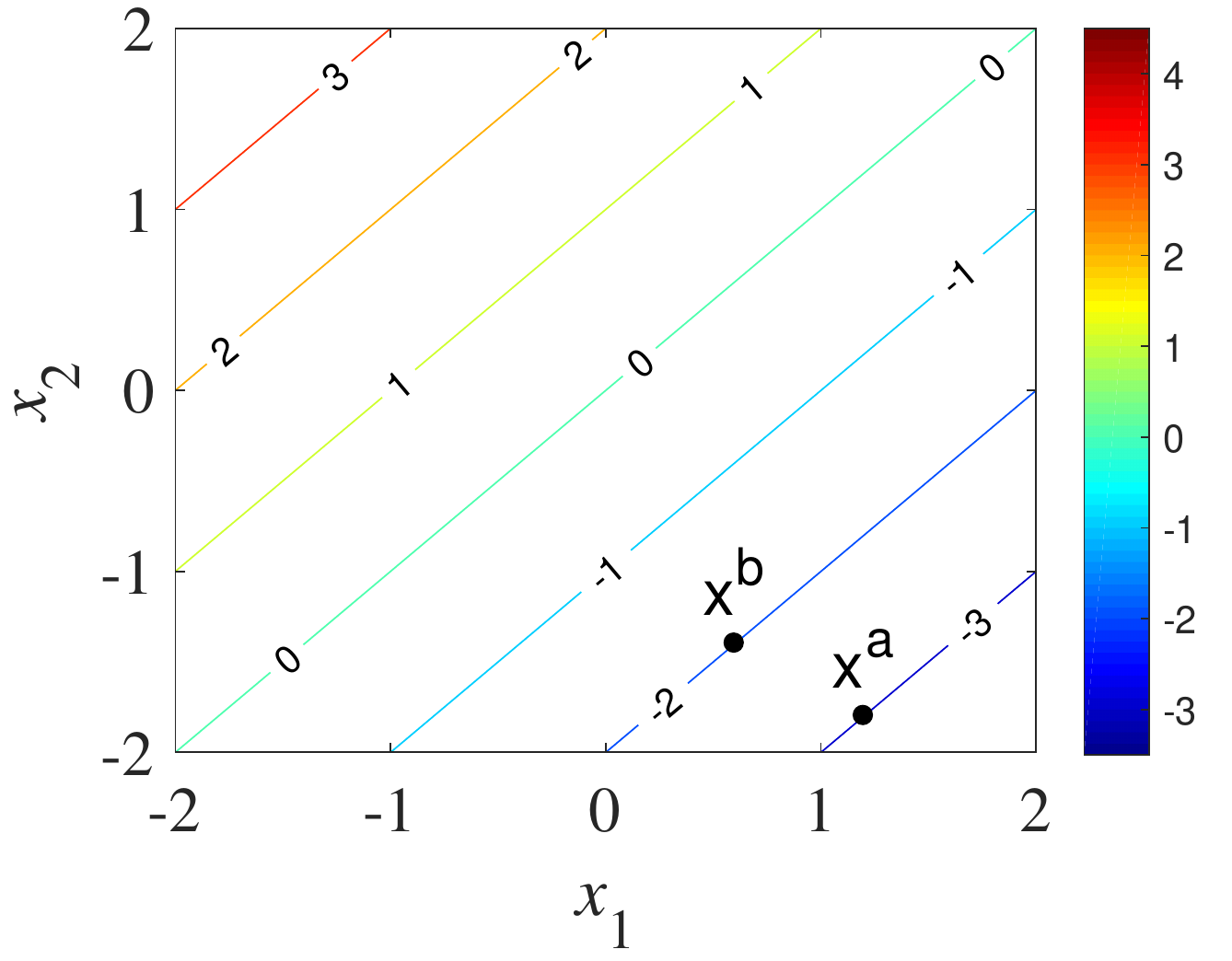}}
\subfigure[\label{Fig:Harmony_example2}The contour of the $j$th constraint function ($f_j$)]{\includegraphics[width=0.32\textwidth]{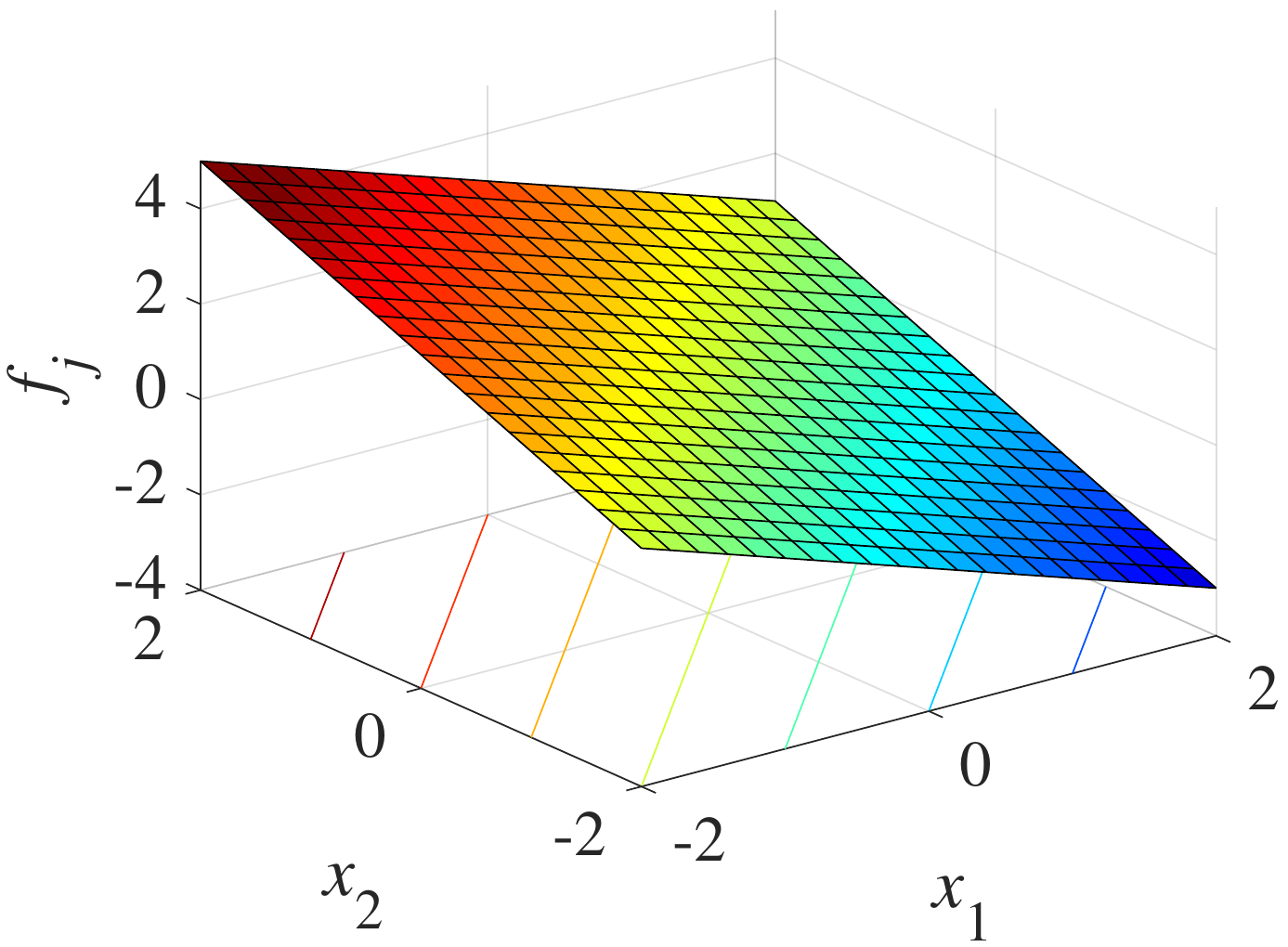}\includegraphics[width=0.07\textwidth]{Arrow}\includegraphics[width=0.32\textwidth]{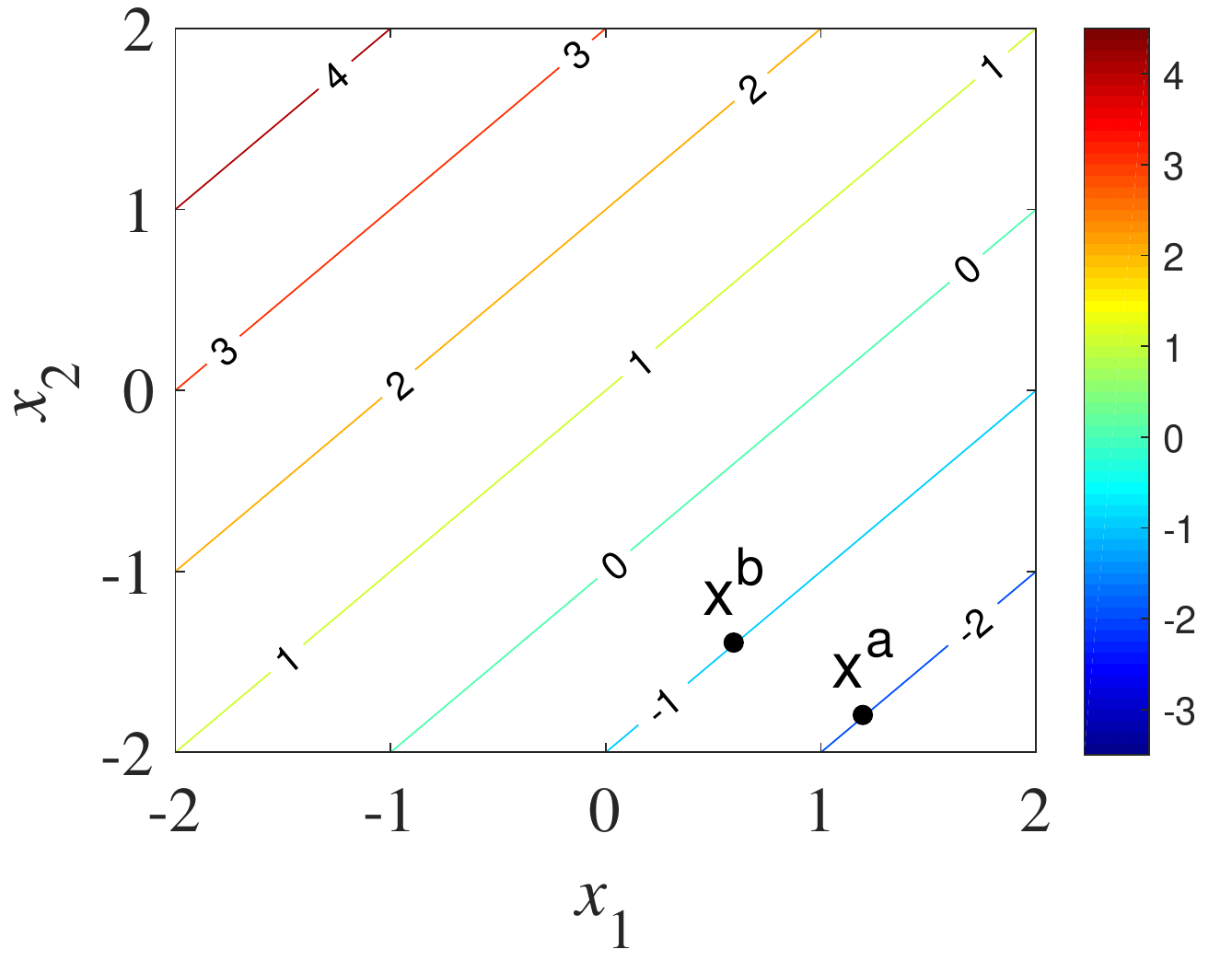}}
\subfigure[\label{Fig:Harmony_example}Evidence of harmony between two individuals]{\includegraphics[width=0.6\textwidth]{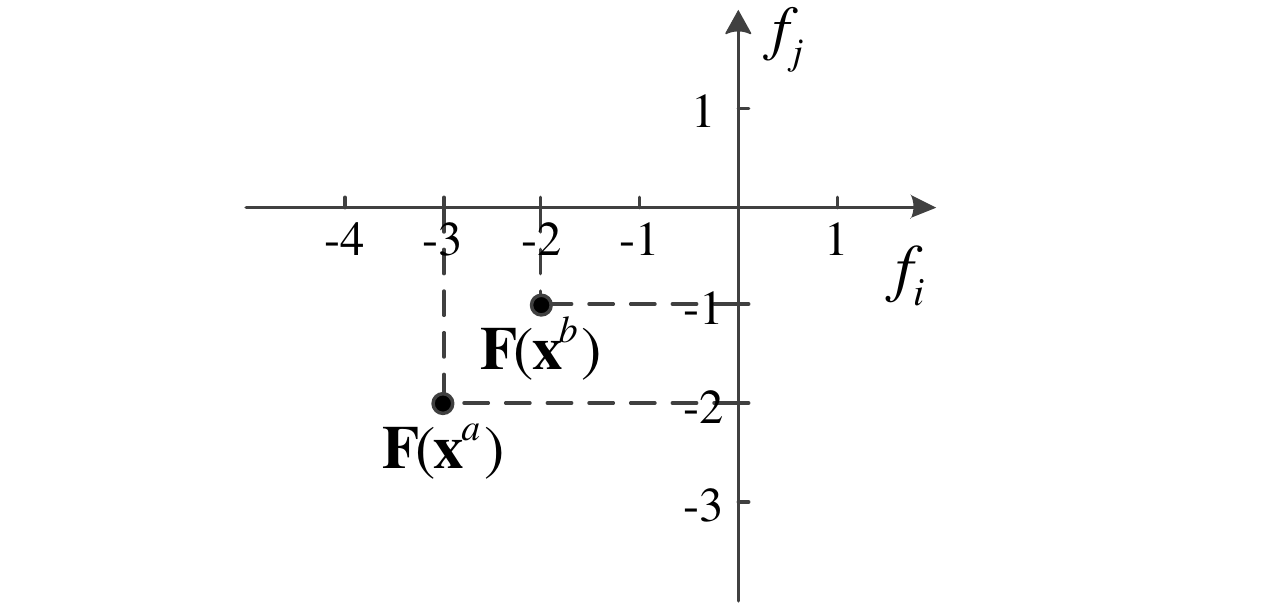}}
\par\end{centering}
\caption{Illustration of two harmonious constraints.}
\label{Fig:Harmony}
\end{figure}

Also, the condition of \textit{total harmony} is illustrated by an example in which constraints $i$ and $j$ are $-x_1+x_2\leq 0$ and $-x_1+x_2+1\leq 0$, respectively, as show in \fref{Fig:Harmony_example1} and \fref{Fig:Harmony_example2}. Their search direction is identical, i.e., towards the bottom right. The relevant $\mathbf{x}^a$-to-$\mathbf{x}^b$, which shows evidence of harmony regarding the two constraints, can refer to \fref{Fig:Harmony_example}.

It can be observed that the feasible region of constraint $j$ is completely included by that of constraint $i$. Once constraint $j$ is satisfied, then constraint $i$ is certainly met, such that constraint $i$ makes little sense in this problem. 
Specifically, for a pair of harmonious constraints, one member whose feasible region covers the other's can be considered redundant unless the decision-maker imposes additional preference.

Accordingly, whether or not to include the redundant constraints remains a question. It is possibly argued that redundant constraints may contain useful information and contribute to better diversity. Yet the advantage it may bring is far less than the benefit from removing redundant constraints. For example, the removing measure can help relieve the computational pressure in both performance evaluations and comparisons. What's more, based on the concepts about Pareto dominance, such a pruning does not affect the partial ordering of the candidate solutions, and thus progress towards the optimal solution is unaffected. 

In this light, removing redundant constraints in the optimization process can be a practical way to eliminate the extra burden on the search under too many constraints. Then the question falls on how to identify the redundant constraints of a problem in-hand.

\subsection{Independent Constraints}
\label{sec:Independence}
In this paper \textit{independence} refers to the relationship wherein two constraints do not affect each other. For example, in a global problem, if there are independent sets of constraints with associated independent sets of decision variables, those sets will have no interplay and can be solved separately.

\begin{figure}[!t]
\begin{centering}
\subfigure[\label{Fig:Independence_example1}The contour of the $i$th constraint function ($f_i$)]{\includegraphics[width=0.32\textwidth]{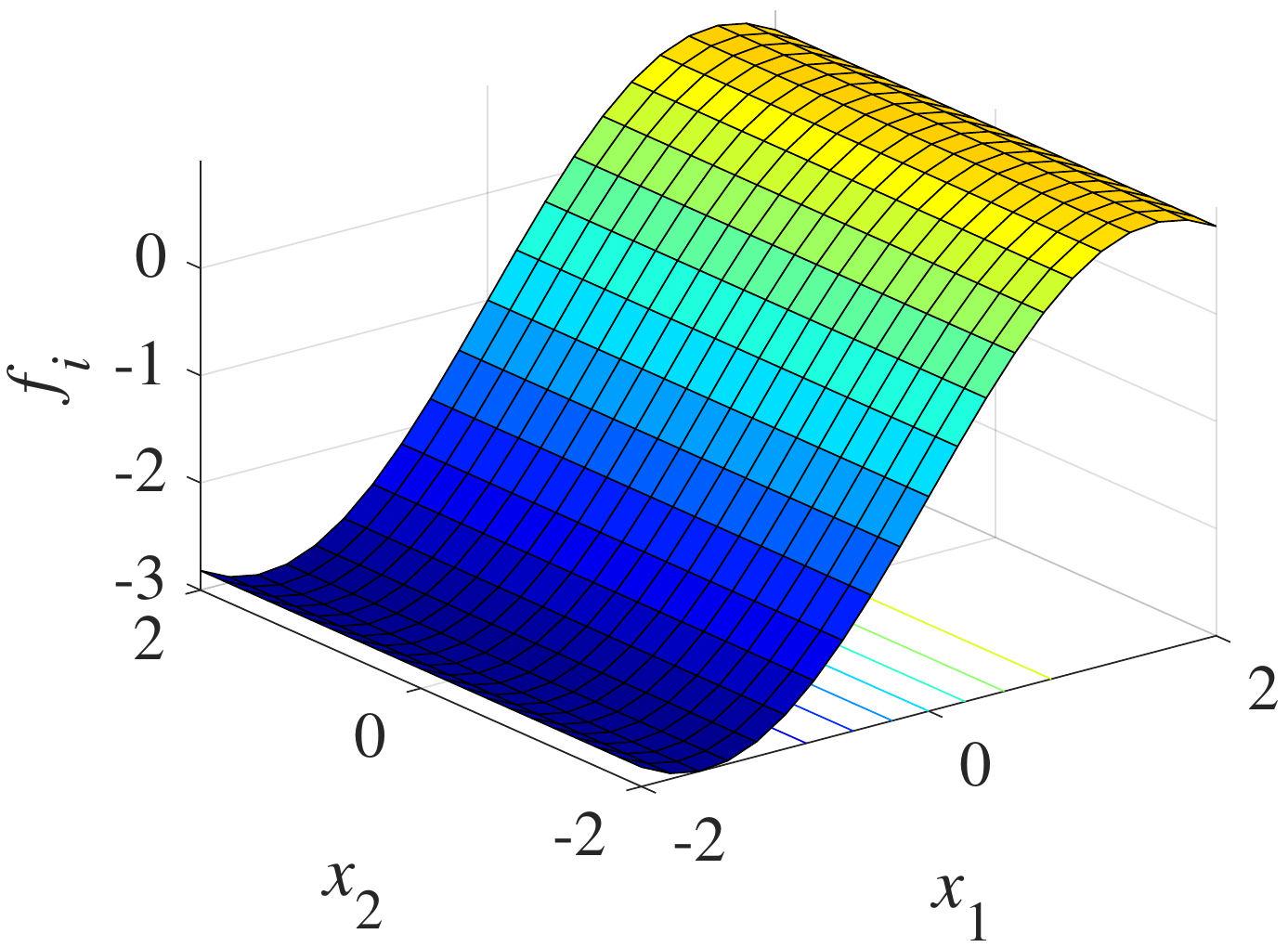}\includegraphics[width=0.07\textwidth]{Arrow}\includegraphics[width=0.32\textwidth]{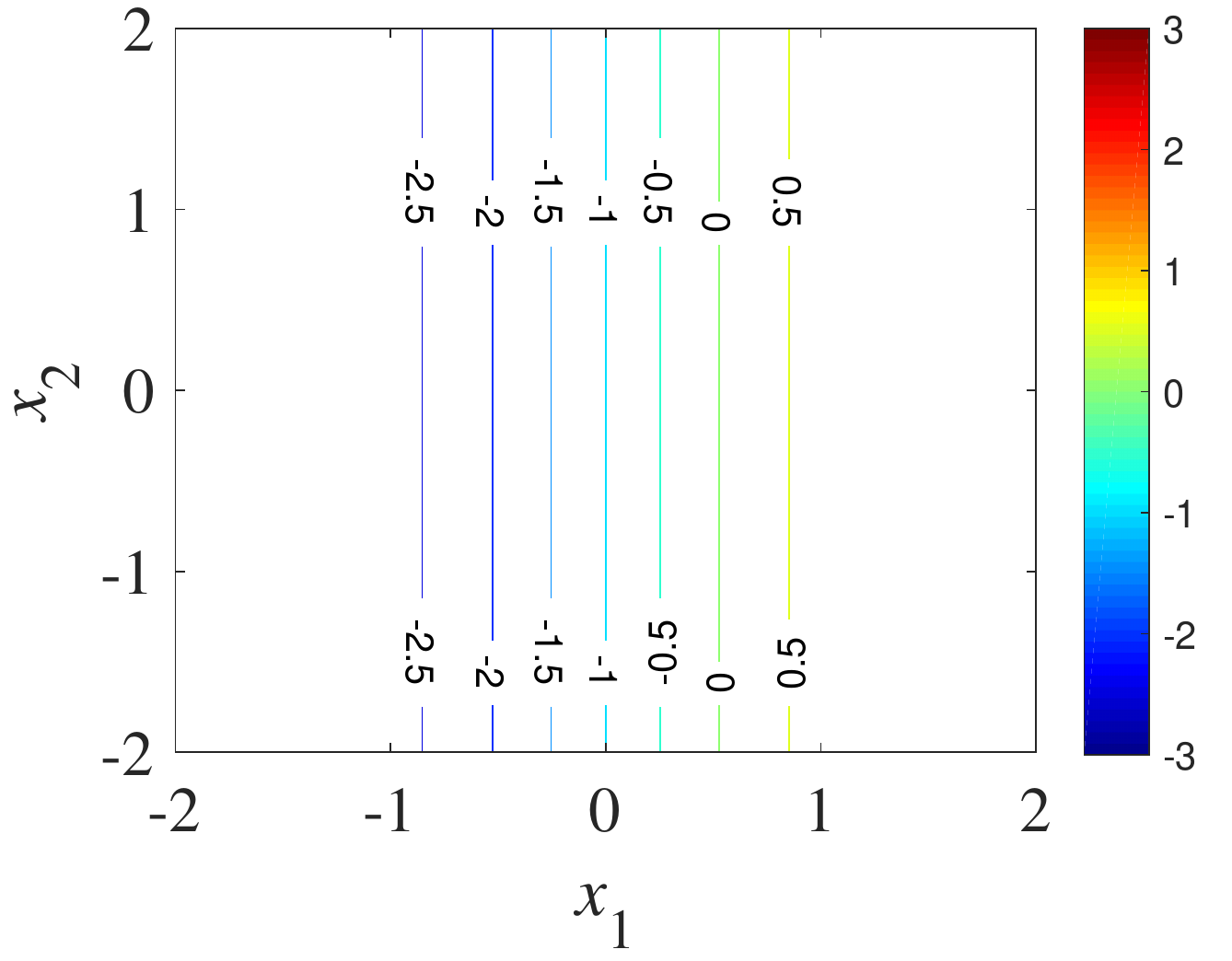}}
\subfigure[\label{Fig:Independence_example2}The contour of the $j$th constraint function ($f_j$)]{\includegraphics[width=0.32\textwidth]{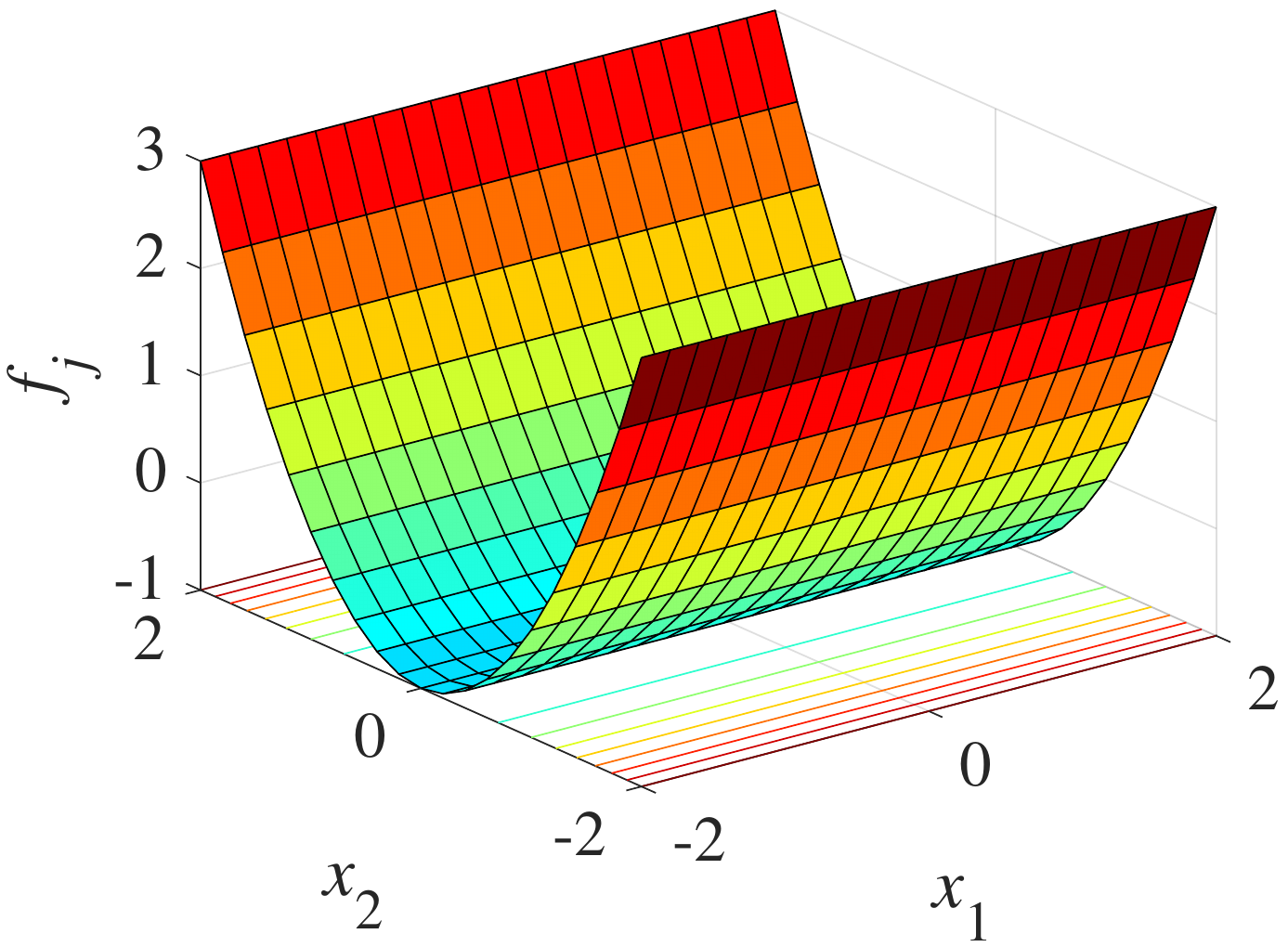}\includegraphics[width=0.07\textwidth]{Arrow}\includegraphics[width=0.32\textwidth]{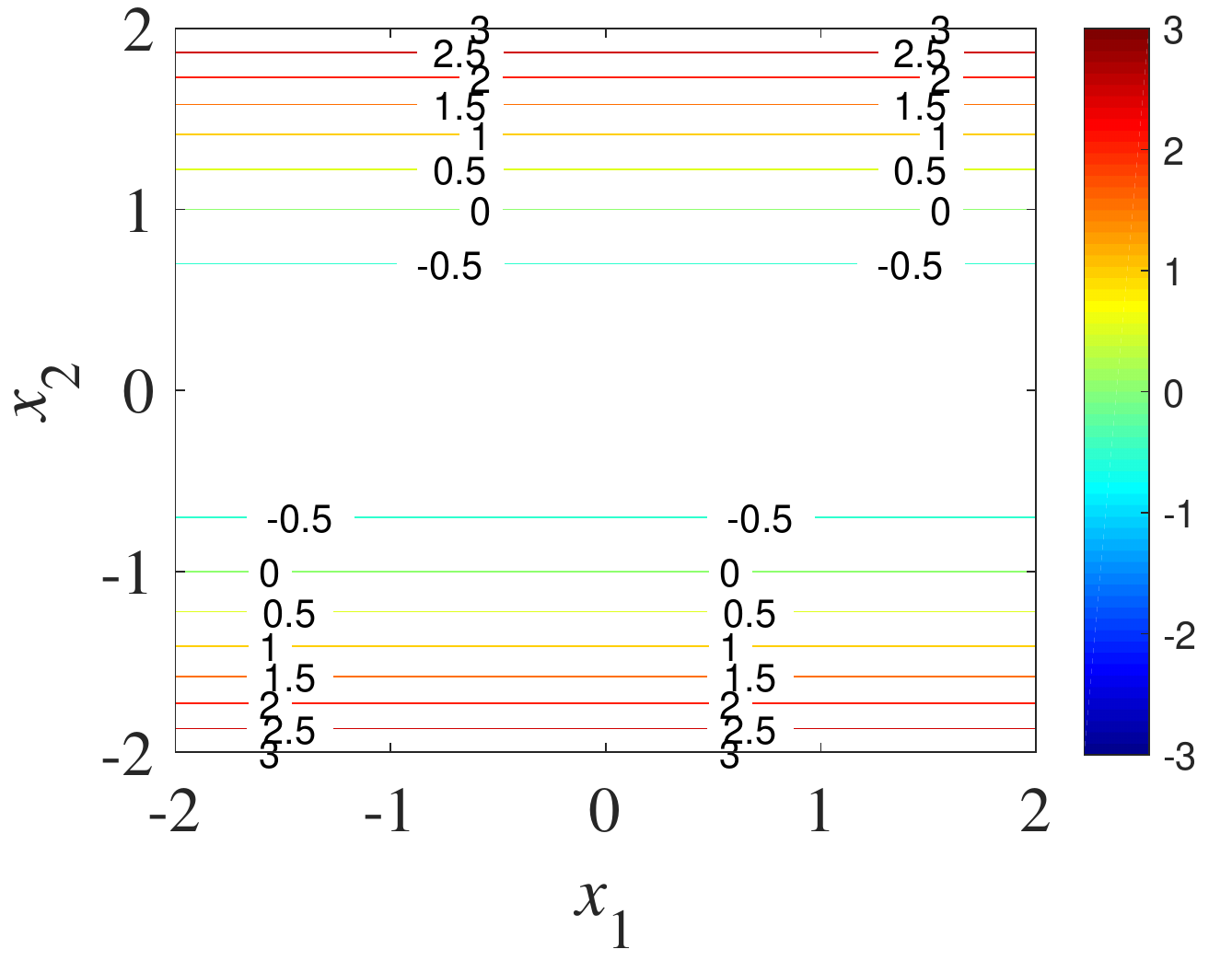}}
\par\end{centering}
\caption{Illustration of two independent constraints.}
\label{Fig:Independence_example}
\end{figure}

The two constraints in \fref{Fig:Independence_example} illustrate the mutual independence between them. Specifically, their mathematical expressions are $2\sin x_1-1\leq 0$ and ${x_2}^2-1\leq 0$ which involve $x_1$ and $x_2$ respectively. It can be seen that the value of $f_i$ depends on the variable $x_1$ while $x_2$ cannot produce any effects on the functional value. Yet the $j$th constraint in \fref{Fig:Independence_example} holds the opposite condition. Thus the change of the functional value of constraint $i$ cannot have any impact on constraint $j$, and vice versa.  

The independent relationship greatly differs from other relationships. In the harmonious and conflicting cases, a single modification for one of the constraints will naturally produce an increase or reduction in the functional value of the counterpart. However, as for independence, the change in one constraint does not affect the other unless we implement appropriate adjustments in distinct parts of the solution.

This relationship implicates the capability to decompose the global optimization problem into a group of sub-problems which optimize separately from each other. 
If advanced knowledge of the independent constraints is available then they can be used to guide the decomposition of the problem. Correspondingly the resources such as candidate solution evaluations are assigned to the optimization of different sub-problems. Given that it is possibly hard to get accurate information about the independent constraints in advance, an online strategy may need to be applied to the population sample data to identify the relationship.

\section{Methods for Identifying Pair-wise Relationships}
\label{sec:Method}
\begin{figure}[tb]
\begin{centering}
\includegraphics[width=0.36\textwidth]{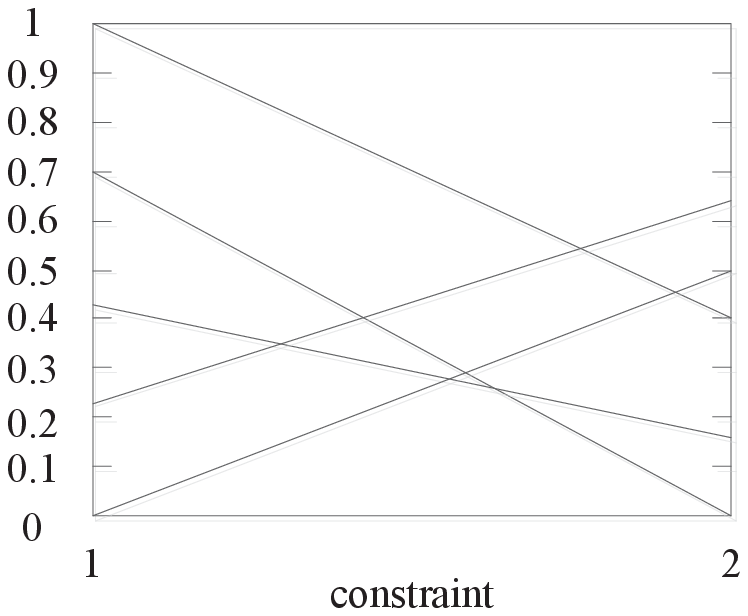}
\par\end{centering}
\caption{Parallel coordinates representation of the constraint values.}
\label{Fig:Parallel_cor}
\end{figure}

The method of parallel coordinates, first described in~\cite{inselberg1985plane}, can be used to present the performance of the constraints by straight lines. Specifically, when investigating a pair of constraints, each solution can obtain its constraint vector which is composed of the functional values of the pair-wise constraints. Then the constraint vector can be represented as a straight line with vertices on the vertical axes where the position of the vertex on the $i$th vertical axis indicates the functional value of $i$th constraint. 
The lines representing two constraint vectors will cross if conflict is exhibited or will fail to cross if harmony is observed. Thus, the magnitude of conflict is visualized as the number of crossed lines, and by contrast, the number of uncrossed lines can reflect the harmony magnitude, as shown in \fref{Fig:Parallel_cor}.

\begin{figure}[tb]
\begin{centering}
\includegraphics[width=0.32\textwidth]{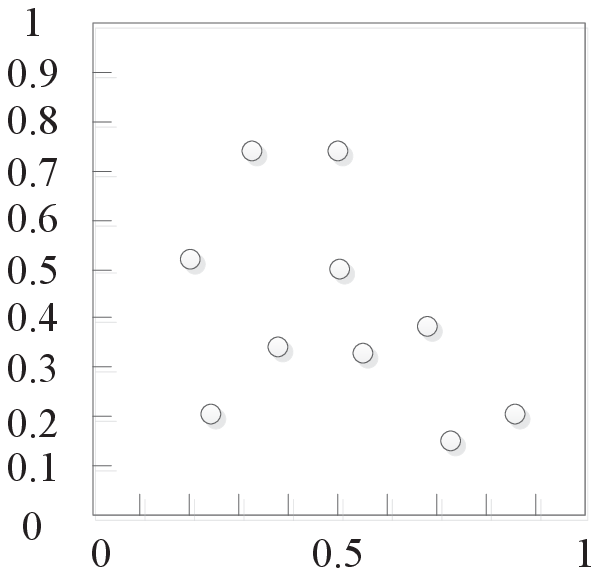}
\par\end{centering}
\caption{Scatterplot representation of the constraint vectors.}
\label{Fig:Scatterplot}
\end{figure}

Another method for identifying the inner relationship is to investigate the Pareto domination relationships~\cite{deb2002nsga} between constraint vectors. Given a pair of constraint vectors, one of them will dominate the other if the corresponding constraints are harmonious, otherwise, they exhibit no domination. Thus for two conflicting constraints, a trade-off surface exists in the constraint-space. As for the harmonious condition, the strength of Pareto dominance can represent the extent of harmony. 
For a set of constraint vector instances, the harmony and the conflict magnitude are able to be reflected by the number of Pareto-dominance pairs and no-dominance pairs, respectively, even though these two relationships co-exist on two constraints, as shown in \fref{Fig:Scatterplot}.

\begin{figure}[tb]
\begin{centering}
\includegraphics[width=0.32\textwidth]{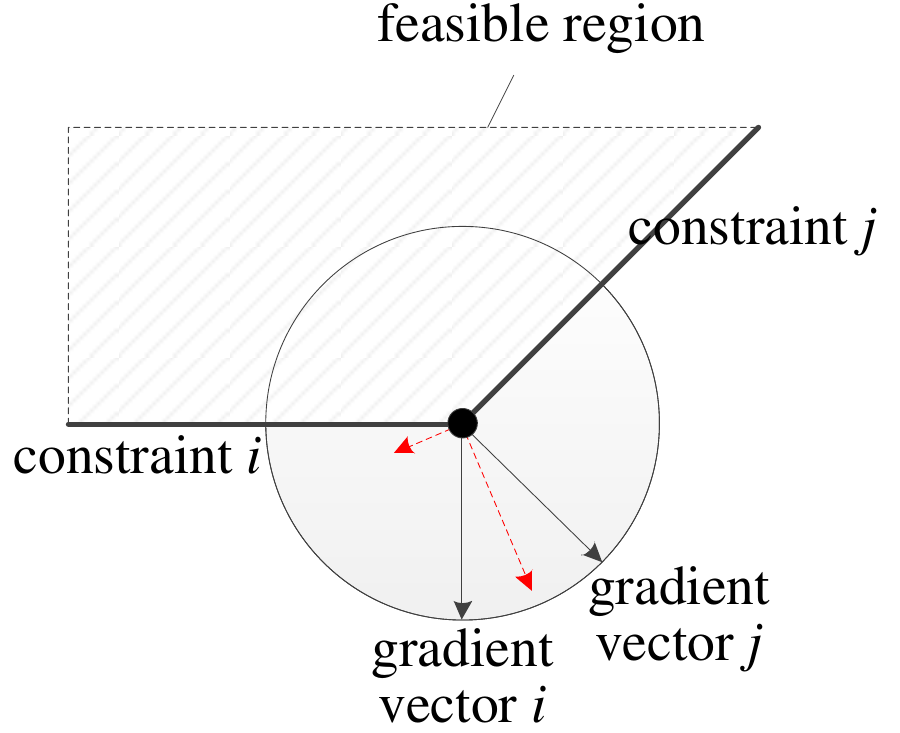}
\par\end{centering}
\caption{Gradient vector representation of the constraints.}
\label{Fig:Gradientvector}
\end{figure}

Furthermore, the gradient vectors, which indicate the incremental direction of the function, can be used to evaluate the relationship between constraints having evident gradient information. Specifically, if the gradient vectors are in the same direction, then the push along a gradient vector will lead to an increase in the functional value of two constraints. On the contrary, the constraints with opposite gradient vectors will produce the inverse change in the functional value during the search. The above two conditions can serve as the judgment criterion for harmony and conflict respectively. In addition, a pair of gradient vectors may form an intersection angle, as shown in \fref{Fig:Gradientvector}. Under this condition, each gradient vector can be decomposed into two sub-vectors, that is, in the direction of the centre line and the perpendicular direction to the centre line. Thus the length of the sub-vectors can represent the magnitude of harmony and conflict, respectively.

\section{Transitivity of Constraint Relationships}
\label{sec:Transitivity}
The pair-wise constraint relationships can be roughly classified into ``harmony'', ``conflict'' and ``independence'' as mentioned above. Correspondingly, whether these constraint relationships can be transitive is to be discussed. 
For example, if the $i$th and the $j$th constraints are harmonious, and so do the $j$th and the $k$th constraints, then the question is whether we can ensure that harmony exists between the $i$th and the $k$th constraints. 

\subsection{Harmony Meets Harmony}
Assume that there are two pairs of harmonious constraints and one in each pair is the same constraint, such as $i$-$j$, and $j$-$k$.
According to the harmony definition, the constraints $i$ and $j$ exhibiting evidence of harmony means that $\exists (\mathbf{x}^a,\mathbf{x}^b)$ satisfies $f_i(\mathbf{x}^a)<f_i(\mathbf{x}^b)$ and $f_j(\mathbf{x}^a)<f_j(\mathbf{x}^b)$. 
Similarly, as for constraints $j$ and $k$, at least two solutions (e.g. $\mathbf{x}^c$ and $\mathbf{x}^d)$ can be found to satisfy $f_j(\mathbf{x}^c)<f_j(\mathbf{x}^d)$ and $f_k(\mathbf{x}^c)<f_k(\mathbf{x}^d)$. 

As the harmony relationship can be total harmony or partial harmony, the two cases are to be discussed separately in the following part. 

First, as a special type of constraint relationship, one characteristic of the total harmony is that any two constraint vector instances are mutually Pareto dominated or dominating. 
That is, $\forall (\mathbf{x}^a,\mathbf{x}^b)$, if $f_i(\mathbf{x}^a)<f_i(\mathbf{x}^b)$, then $f_j(\mathbf{x}^a)<f_j(\mathbf{x}^b)$. The principle goes for another pair of constraints as well. Specifically, $\forall (\mathbf{x}^c,\mathbf{x}^d)$, if $f_j(\mathbf{x}^c)<f_j(\mathbf{x}^d)$, then $f_k(\mathbf{x}^c)<f_k(\mathbf{x}^d)$. Note that ($\mathbf{x}^c,\mathbf{x}^d$) here can be any solutions. Thus, given any two solutions such as $\mathbf{x}^a$ and $\mathbf{x}^b$, if $f_i(\mathbf{x}^a)<f_i(\mathbf{x}^b)$, then $f_j(\mathbf{x}^a)<f_j(\mathbf{x}^b)$, and then $f_k(\mathbf{x}^a)<f_k(\mathbf{x}^b)$. It can be observed that constraints $i$ and $k$ also holds the condition for total harmony. In other words, the constraint relationship of total harmony can be transitive.

Second, in partial harmonious constraints, not all solution instances are expected to meet the harmony requirements. For example, for constraint vectors $\mathbf{x}^a$ and $\mathbf{x}^b$, even though $f_i(\mathbf{x}^a)<f_i(\mathbf{x}^b)$ and $f_j(\mathbf{x}^a)<f_j(\mathbf{x}^b)$, $f_k(\mathbf{x}^a)$ can be larger or smaller than $f_k(\mathbf{x}^b)$. Thus any kind of relationship can appear for constraints $i$ and $k$. Under this circumstance, the two pairs of constraints show no explicit correlation. 

\subsection{Harmony Meets Conflict}
Similar to the analysis in the last subsection, first we concentrate on total harmony in constraints $i$-$j$ and total conflict in constraints $j$-$k$.
According to the definition, the following conditions are required to be satisfied:
\begin{equation}
\begin{gathered}
\forall(\mathbf{x}^a,\mathbf{x}^b), \text{if} f_i(\mathbf{x}^a)<f_i(\mathbf{x}^b), \text{then} f_j(\mathbf{x}^a)<f_j(\mathbf{x}^b);
\\
\forall(\mathbf{x}^a,\mathbf{x}^b), \text{if} f_j(\mathbf{x}^a)<f_j(\mathbf{x}^b), \text{then} f_k(\mathbf{x}^a)>f_k(\mathbf{x}^b).
\notag
\end{gathered}
\end{equation}
Based on the above conditions, it can be obtained that 
\begin{equation}
\begin{gathered}
\forall(\mathbf{x}^a,\mathbf{x}^b), \text{if} f_i(\mathbf{x}^a)<f_i(\mathbf{x}^b), \text{then} f_k(\mathbf{x}^a)>f_k(\mathbf{x}^b).
\notag
\end{gathered}
\end{equation}
This means that the constraints $i$ and $k$ are in total conflict if $i$-$j$ constraints are total harmonious and $j$-$k$ constraints are total conflicting.

However, when any of the two conditions are not met, the conclusion cannot be obtained, that is, the transitivity between harmony and conflict only appears from the total ones.

\subsection{Conflict Meets Conflict}
When a constraint (e.g., constraint $j$) is totally conflicting with the other in two pairs of constraints (e.g., $i$-$j$ and $j$-$k$), what relationship can be produced between constraints $i$ and $k$ is first discussed.

Assume that the two pairs of totally conflicting relationships are expressed as follows:
\begin{equation}
\begin{gathered}
\forall(\mathbf{x}^a,\mathbf{x}^b), \text{if} f_i(\mathbf{x}^a)<f_i(\mathbf{x}^b), \text{then} f_j(\mathbf{x}^a)>f_j(\mathbf{x}^b);
\\
\forall(\mathbf{x}^a,\mathbf{x}^b), \text{if} f_j(\mathbf{x}^a)<f_j(\mathbf{x}^b), \text{then} f_k(\mathbf{x}^a)>f_k(\mathbf{x}^b).
\notag
\end{gathered}
\end{equation}
Given a pair of constraint vectors, e.g., $\mathbf{x}^a$ and $\mathbf{x}^b$, according to the first condition, if $f_i(\mathbf{x}^a)<f_i(\mathbf{x}^b)$, then $f_j(\mathbf{x}^a)>f_j(\mathbf{x}^b)$, that is actually $f_j(\mathbf{x}^b)<f_j(\mathbf{x}^a)$. Also, according to the second condition, it can be obtained that $f_k(\mathbf{x}^b)>f_k(\mathbf{x}^a)$, which can be converted as $f_k(\mathbf{x}^a)<f_k(\mathbf{x}^b)$. In sum, we conclude that $\forall(\mathbf{x}^a,\mathbf{x}^b)$, if $f_i(\mathbf{x}^a)<f_i(\mathbf{x}^b)$, then $f_k(\mathbf{x}^a)<f_k(\mathbf{x}^b)$. In other words, constraints $i$ and $k$ are total harmonious when two associated pairs of constraints are in total conflict.

In addition, if the conflicting relationship is just partial conflict, then the above conditions cannot be guaranteed to be true, and the transitivity no longer holds.

\subsection{Independence Meets Other Relationships}
The independent relationship essentially indicates that the two constraints have independent variables or do not affect each other. Herein constraints $i$ and $j$ are assumed to be independent. 
With respect to another pair of constraints, i.e., $j$ and $k$, they can be harmonious, conflicting, or independent. No matter in which relationship, it is unknown that what variables are involved in the constraint $k$ and whether constraints $i$ and $k$ will affect the other's functional value. 
Thus, it is hard to determine the relationship between $i$ and $k$ just based on known relationships. 

\section{Conclusion}\label{sec:conclusion}
Many engineering problems involve in handling a large number of constraints. However, few studies in the literature have explicitly investigated the constraint relationships. To facilitate the treatment of large number of constraints in future work, this study, to the best of the authors' knowledge, has made the first initial analysis of pair-wise constraint relationships. Specifically, the conflict, harmony, and independent relationships are  identified, motivated by the comparison of functional value in constraints. Amongst them, the conflicting constraints pose a challenge to search for optimal feasible solutions.  As for harmonious constraints, they do not have a severe impact on the problem as do conflicting ones, and it is recommended to prune the redundant constraint(s) if they are total harmonious. Besides, independent relationship enables the problem to be decomposed into multiple sub-problems. Then three methods are summarized to identify the pair-wise relationship, with the magnitude of harmony and conflict evaluated. Moreover, whether the pair-wise relationship can be transitive or not is further discussed, providing a baseline in grouping constraints.

This study is limited in a certain number of ways. First, the analysis is mainly presented in a qualitative aspect. The accurate quantitative evaluation is further needed. Second, how to  identify constraint relationships in an online manner remains challenging. The future research will mainly embark on the above two aspects.


%







\bibliographystyle{splncs04}
\bibliography{bibAllpaper}

%

%








\end{document}